\documentclass{article}

\PassOptionsToPackage{numbers, compress}{natbib}
\usepackage[final]{neurips_2024}
\usepackage{times}
\usepackage{amsmath}
\usepackage{amssymb}
\usepackage{multicol,float}
\usepackage{vwcol}  
\usepackage{graphicx}
\usepackage{url}

\newcommand{\meta}{\texttt{Meta}}
\newcommand{\commonvoice}{\texttt{CommonVoice}}
\newcommand{\sonos}{\texttt{Sonos}}

\newcommand{\wtvt}{\texttt{W2V2}}

\newcommand{\wavlm}{\texttt{WavLM}}

\newcommand{\largem}{\texttt{large}}
\newcommand{\basem}{\texttt{base}}
\usepackage[utf8]{inputenc} % allow utf-8 input
\usepackage[T1]{fontenc}    % use 8-bit T1 fonts
\usepackage{hyperref}       % hyperlinks
\usepackage{url}            % simple URL typesetting
\usepackage{booktabs}       % professional-quality tables
\usepackage{amsfonts}       % blackboard math symbols
\usepackage{nicefrac}       % compact symbols for 1/2, etc.
\usepackage{microtype}      % microtypography
\usepackage{xcolor}         % colors
\usepackage{xpatch}

\usepackage{xpatch}
\makeatletter
\xapptocmd{\NAT@bibsetnum}{\setlength{\leftmargin}{0pt}\setlength{\itemindent}{\labelwidth}\addtolength{\itemindent}{\labelsep}}{}{}
\makeatother

\title{Speaker Group Encoding in Self-supervised Speech Recognition Models}

\author{%
  Felix Herron$^{1,2}$\And Solange Rossato$^{2}$ \And Alexandre Allauzen$^{1}$ \And Benoit Favre$^{2,3}$ \And François Portet$^{2}$ \vspace{0.5cm}\\
    MILES Team, LAMSADE, Université Paris Dauphine-PSL, France (1)\\
    GETALP Team, LIG, Université Grenoble Alpes, France (2)\\
    NLP team, LIS, Aix-Marseille University, France (3)\\
  \texttt{felix.herron@daupine.eu} \\
}

\begin{document}

\maketitle

\begin{abstract}
    We investigate what self-supervised speech recognition models (S3Ms) learn about speaker groups (SGs). We examine several states of S3Ms: pretrained, finetuned on speaker identification (SID), finetuned on automatic speech recognition (ASR), and ASR-finetuned using a fairness enhancing algorithm. We find that S3Ms encode information about several speaker group categories (SGCs), including their gender, age, dialect, ethnicity, and whether they are a native speaker. We find that finetuning for SID amplifies certain SGCs, namely those whose variance is more phonetic in nature, though it does not amplify other SGCs, namely those whose variance is more semantic in nature. On the other hand, finetuning for ASR discards phonetically variant speaker group information (SGI) but retains semantically variant SGI. We find that ASR algorithms designed for fairness improvement change to what extent SGI is encoded in S3Ms; however, this is primarily true for for phonetically variant SGCs, and less true for semantically variant SGCs. We discuss how SGI is encoded by each layer, and identify subdimensions of embeddings responsible for encoding different SGCs. Finally, we discuss how our findings could be beneficial in designing fairer ASR algorithms.

\end{abstract}% TSD 2025:
% TSD 2025: keywords, comma-separated

% \section{TODOs}
% \begin{enumerate}
%     \item Finish related work (1h)
%     \item In-depth on probing technique (aha)
%     \item Redo flow chart final version with
%     \begin{itemize}
%         \item One version with pretrained vs ASR vs SID
%         \item One verison with ASR vs DAT/DET/DAT-DET
%     \end{itemize}
%     \item Dump log_reg_comp model, redo weights analysis (scatter/grid/entropy for each category)
%     \item Conclude: models that learn SID learn SGI ? Fairer models can better differentiate between SGs ?
%     \item (Freeze feature extractor, learn SG classifier ?)
% \end{enumerate}

\section{Introduction}

Self-supervised speech encoder models (S3Ms) are popular backbone tools for many speech-related downstream tasks. They are particularly versatile because they learn task-agnostic embeddings during pretraining, which requires pure audio data only. The rich embeddings learned during pretraining can be paired with lightweight downstream models to finetune an S3M to specific downstream tasks such as automatic speech recognition (ASR) or speaker identification (SID), among many others \cite{superb}. The lightweight downstream models learn to map from the latent output space of S3Ms to a downstream task space.

There has been considerable study showing that S3Ms perform better on downstream tasks for certain speaker groups (SGs) than others. For example, non-native speakers are shown to be understood less well by ASR systems \cite{towards, mandarin_non-native_asr, sonos}; the same is true of non-standard dialects of various languages \cite{towards, sonos, survey_on_accents_det_dat, portuguese_regional, survey_on_accents_det_dat}, as well as very young and very old speakers \cite{towards, sonos}. These observations support the hypothesis that S3Ms model utterances of different SGs differently. This paper addresses the question as to how these differences are manifested in an S3M's layer-by-layer processing of an utterance. We present analyses of which types of speaker group information (SGI) are present at each layer for various model architectures, and how this SGI presence shifts as models are finetuned. We also analyze the neuron-level representations of SGI, and discuss how SGI is distributed across latent embedding dimensions. Our paper is focused on interpretability of S3Ms, and while we don't provide a concrete fairness promoting algorithm based on our work, we suggest potential avenues for future work in fairness promotion based on our results.

%For example, do S3Ms model men's speech differently than women's, or old people differently than young people? 

% Specifically, we are curious to what extent certain measurable SG characteristics are implicitly encoded by S3M models. For example, do S3Ms model men differently than women, or old people differently than young people?  Our analysis consists of linear probing experiments using speaker embeddings extracted from different locations in various S3M models. We examine to what extent S3M embeddings contain information corresponding to SG characteristics (such as gender, dialect, age, etc.), and which types of models excel at this. Finally, we study the relationship between SGI capture and downstream task performance.

%shows inferior ASR performance on non-native speakers and some regional accents, as well as children. \cite{amazon_kmeans} shows bias against speakers based on their geographic location. This research tends to focus either on identifying a lack of fairness \cite{towards, amazon_kmeans, mandarin_non-native_asr, arabic_asr, portuguese_regional, solange_bias_women}, or proposing models that are fairer \cite{mtl_dat_comp, aave_separate_models, survey_on_accents_det_dat, dasBestBothWorlds2021, amazon_kmeans, towards, vc_article}

\section{Related work}

\subsection{S3M interpretability}

The nature of transformer-based speech S3M embeddings has been studied since their inception in 2021 \cite{wav2vec2}. For example, they are used in benchmarks such as SUPERB in English or LeBenchmark in French, where they have been shown able to solve any number of speech related tasks with or without finetuning \cite{superb, lebenchmark}. In \cite{wavlm}, authors analyze which layers perform best on a variety of downstream tasks, showing that earlier layers tend to contain more information on speaker identity related tasks (such as SID or speaker diarization), while later layers are better suited for more semantic tasks (such as ASR or intent classification). Building on this, \cite{toyota_layerwise_interpretability} and \cite{acoustic_word_embeddings_goldwater} show that phonetic information peaks in middle layers and word identity information in later layers. Furthermore, they showed S3Ms can be viewed as autoencoders, with later layers learning features similar to initial layers (logical given reconstructive S3M pretraining objectives, generally cross-entropy or contrastive loss for masked input subsequences \cite{bestrq_ryan}), though this trend disappears upon ASR finetuning. \cite{toyota_layerwise_interpretability_2} shows that S3Ms which use CNNs as acoustic feature extractors (rather than using Mel filterbank coefficients (MFCs), for example), tend to learn a representation similar to MFCs anyway. They also show that, depending on pretraining objective, different models store phone/word information in different layers. \cite{phonetics_over_semantics_in_s3m} shows that S3Ms encode some semantic information of word meaning, though significantly less than they do raw phonetic content.

There has also been work on speaker-level layerwise interpretability. For example, \cite{adversarial_and_enhancing} shows that S3Ms finetuned for ASR achieve high SID performance for early layers, but lose it almost entirely in final layers. Furthermore, \cite{orthogonality_first} shows that speaker information is stored orthogonally from phoneme information in intermediate S3M representations, and \cite{crv_pca} provides a method to quantitatively compare the orthogonality of types of information (such as speaker information vs phoneme information). \cite{silence_ssl_hubert} shows that the silent portions of S3M embeddings contain more speaker information than the sections with actual speech, and that adding silence artificially to an utterance boosts an S3M's ability to model the speaker of that utterance. 

We emphasize one key difference between SID vs SG identification (SGID) - the former involves learning to recognize a fixed set of speakers, while the latter must generalize to unknown speakers. With this in mind, high performing SID does not necessarily generalize to high performing SGID. There has been comparatively little work in SGID - many publicly available speech corpora contain either few or inconsistently labeled metadata categories, so it is challenging to study. One case where there has been some work is in Arabic dialect identification \cite{arabic_dialect_id}. This work is thus somewhat related to ours, though there is a far greater distance between different dialects of Arabic than dialects of English, and this distance is often heavily word-based as opposed to phonetic \cite{similarities_arabic_dialects}.

\section{Methods}
\label{sec:methods}

% \begin{figure}[t]
%   \centering
%   \includegraphics[width=\linewidth]{img/pipeline.png}
%   \caption{Speaker group (SG) metadata probing pipeline}
%   \label{fig:pipeline}
% \end{figure}

%During finetuning, the weights of the S3M are often slightly tweaked to conform to the specific downstream task; however, the bulk of learning is assumed to occur during pretraining.

% Pretraining takes place by leveraging a . 

An S3M processes an utterance as a sequence of discretized acoustic embeddings corresponding to its length, and produces a corresponding sequence of rich embeddings, one for each discretized unit. \cite{mean_pooling_sid} show that taking the mean over all embeddings of an utterance produces a fixed-length vector that contains information related to the speaker's identity. This is logical, as an average over all embeddings should cancel out local phonetic information and amplify global utterance-level information. We experiment with several pooling strategies, similarly to \cite{silence_ssl_hubert}, by taking the mean over a subsequence of frame embeddings. We experiment with five subsequences: the first frame, first 50 frames (corresponding to one second of audio), final frame, final 50 frames (i.e. the final second of audio), and the entire sequence. We calculate these pooled embeddings for each model at each layer.

In order to test for SGI, we first experimented with linear probes, as was proposed in the SUPERB benchmark for the SID task \cite{superb}. These probes take the form of single linear layers: $M_{SGC} \in \mathbb{R}^{d_{M}, |SGC|}$, where $d_M$ is the size of each model embedding (i.e. 1024 for the \largem{} S3M configuration), and $|SGC|$ is the number of classes in the SGC (i.e. 2 for gender\footnote{For the sake of simplicity, we assume two genders throughout this paper.}). However, we noticed that our probes tended to overfit on the speakers in our training set, resulting in huge performance differences in SGID between our train and test sets. (This is the manifestation of the effect of the fundamental difference between SGI and SID!). Our probes were not only learning general SG features, they were learning to map specific speakers to SG classes. To counteract this, we split our probe into two layers (see Equation \ref{eq:lin_reg}): one projection layer $M_P$ of small latent dimensionality $d_P$, i.e. 5, and one classification layer $M_C$, with $M_P \in \mathbb{R}^{d_M, d_P}, M_C \in \mathbb{R}^{d_P, |SGC|}$. We then learn two linear connections leading from this layer: 1) the SG classification layer as before, and 2) a speaker identification layer, which we train using a reverse gradient layer \cite{dat_first}. This forces the intermediate layers to be agnostic to speaker identification and prevents speaker overfitting. We first warm up all three layers until convergence (without propagating the reverse gradient to $M_P$), then add the gradient reversal and train until a second convergence, as described in \cite{adversarial_and_enhancing}.

The low-dimensional, intermediate embeddings from the projection layer contain embeddings from which a SG classifier can be learned; thus, these intermediate embeddings can be viewed as a projection onto the $SGC$ space. This two-tiered probe does not add any complexity to our probing procedure, as we use no non-linear activation function between $M_P$ and $M_C$\footnote{We experimented with two-layer MLPs with ReLU activation and found no significant improvement in SGI detection}.

\begin{align}
    &\text{M}_{SGC}(u) := M_C M_{P}(u) \quad &(\in \mathbb{R}^{|SGC|}) \label{eq:lin_reg} \\
    &\text{M}_{\text{SID}, SGC}(u_1, u_2) := M_{SID} M_{P}(u) \quad &(\in \mathbb{R}^{|\text{speakers}|}) \label{eq:rev_grad}
\end{align}

In the experiments presented in this paper, we used $d_P=5$, though we experimented with $5 \in [5,10,50]$ and found similar results.

\subsection{Self-supervised speech encoder models (S3Ms)}

We analyzed the capacity of two main S3M architectures, \texttt{Wav2Vec 2.0} (\wtvt{}) and \wavlm{}. We chose completely open-source checkpoints of these models, which we can therefore compare according to their architecture and training data. We used \textbf{\wtvt{}} as it was the first to use the transformer architecture in S3M speech modeling and is the most widely used today \cite{wav2vec2, bestrq_ryan}. Furthermore, the open-weight accessibility of many configurations of \wtvt{} on Hugging Face greatly facilitated our analysis \cite{wav2vec2}. We also analyzed \textbf{\wavlm{}}. Not only is it by far the best open-source performer on the SUPERB benchmark; its pretraining lends it to be a good a priori candidate for encoding SGI. First, during pretraining it uses a reconstruction loss rather than the contrastive loss of \wtvt{}, as inspired by HuBERT \cite{hubert}. This requirement to be able to reproduce fundamental information about the initial signal might make \wavlm{} more likely to store SGI. (Previous work has shown pretraining objective to play an important role in the information captured by S3Ms \cite{ssl_pre_w2v2_layer_analysis}). Secondly, \wavlm{} uses an additional utterance mixing step in pretraining which is meant to enhance embeddings with speaker-specific information \cite{wavlm}. \wavlm{} pretraining involves randomly overlaying unrelated audio during to simulate a noisy speaking environment, so the model must learn how to differentiate between different audio sources and thus might learn to better model SGs.

We perform most of our experiments on the \textbf{\largem{}} configuration of S3Ms. This allows for standardization of model size (24 hidden layers, hidden layer size 1024, $\sim 300m$ trainable parameters). We also experimented with the \basem{} configuration of \wavlm{} ($\sim 100m$ parameters). This will allow us to ascertain whether \largem{} model size is prerequisite for modeling SGI.

%\footnote{We performed complementary analysis on \textbf{base} model configurations, which exhibited similar behavior with slightly lessened ability to recognize SGs. This is in line with the inferior performance generally observed in \textbf{base} models \cite{superb}}.

For maximum comparability, we focused on models trained on a version of LibriVox (LibriLight, VoxPopuli), which is the case for the \wavlm{} models and \wtvt{}-lv60. However, we also compare the 53-language \wtvt{}-XLSR-53. This will allow us to evaluate the importance of diverse pretraining data in some SGID tasks.

\subsection{Finetuning}

Along with model architecture, we also varied our experiments between models that were either \textbf{pretrained only} or \textbf{pretrained and finetuned}. We investigate whether the process of finetuning on ASR or SID leads models to store SGI or the opposite. We also experimented with the Domain Adversarial Training (DAT) and Domain Enhancing Training (DET) finetuning variants, which have been known to improve overall transcription performance \cite{adversarial_and_enhancing} as well as fairness \cite{survey_on_accents_det_dat}. DET training works by appending an additional SG classifier to any given transformer layer (usually a middle layer) to force the middle layer to retain speaker information during finetuning. DAT training works in the opposite manner, by placing an SG classifier on a late layer with a reverse gradient update rule, to force the model to learn speaker invariant representations prior to ASR. This is the same strategy as we used for training speaker-invariant probes in the previous section.

In accordance with \cite{adversarial_and_enhancing} we used speaker ids as classes for our multi-class objective, with DET classifier on the 10th layer and DAT classifier on the 21st for \largem{} models, and 5th and 9th respectively for \basem{}. For our SID classifiers, we train an xvector \cite{xvectors} based on the activations of each layer, an architecture that has been shown to capture speaker information efficiently\footnote{We also experimented with linear classifiers but found this less effective in erasing SGI in adversarial layers. This is likely due to speaker identity in embedding space being of greater dimension than one, and thus a single linear adversarial classifier cannot erase it entirely \cite{linear_adversarial_concept_erasure}}. For ASR, we used a three layer MLP and CTC loss. We used finetuning recipes provided by the SpeechBrain toolkit \cite{speechbrain}.

We finetuned our models using a subset of the \commonvoice{} 16 English dataset. We chose this dataset due its large size and diversity of speakers, and availability of speaker IDs. One potential downside of this choice is that the datasets we use for SGI detection are from smart speaker commands, which is a different type of speech and thus potentially less than perfectly compatible with our finetuned models.

\subsection{Speaker group data}
\label{sec:datasets}

To evaluate models' detection of SGI, we used two datasets designed for researching fairness in ASR. Both the \texttt{Meta Fair-speech} \cite{meta_fair} and \texttt{Sonos Voice Control Bias Assessment} \cite{sonos} datasets are based on recordings of paid amateur participants performing smart speaker commands lasting several seconds. Each command is annotated with precise speaker metadata. \sonos{} contains 170K utterances and is annotated with \textit{gender} (2 classes\footnote{Male; Female}), \textit{dialect} (8 classes\footnote{6 regional dialects within the USA: Inland-north; Mid-Atlantic; Midland; New England; Southern; Western. Also non-native speakers of \textit{either} an Asian language \textit{or} Spanish.}), and speaker \textit{age} (5 classes)\footnote{9-16; 17-28; 29-41; 42-54; 55-100.}. It also contains speaker IDs (1038 speakers in aggregate over train, valid, and test sets) which we used to compare the difficulty of SID vs SGI probing. \meta{} contains 26.5K utterances and is annotated with \textit{gender} (2 classes\footnote{Male; female.}), \textit{ethnicity} (7 classes\footnote{Asian, South Asian or Asian American; Black or African American; Hispanic, Latino or Spanish; Middle Eastern or North African; Native American, American Indian, or Alaska Native; Native Hawaiian or Other Pacific Islander; White.}), \textit{socio-economic background} (3 classes\footnote{Low, Medium, Affluent}), speaker \textit{age} (4 classes\footnote{18-22; 23-30; 31-45; 46-65}), and \textit{native speaker} (2 classes). We used the train-test split provided by \sonos{} for training our SGI probes; we created our own split for \meta{} as at contains none\footnote{\meta{} also doesn't contain speaker IDs; in order to avoid data leakage of having the same speaker in both train and test splits, we created pseudo-ids based on unique combinations of all speaker attributes. However, this was not as fine-grained a split as possible, as we ended up with far fewer pseudo-ids (146) as \meta{} reported as appearing in the corpus (593).}.

\section{Results}

% TODO:
% \begin{itemize}
%     \item Flow plot for SGI for F1/siamese acc
%     \item Number of frames bar plot
%     \item Grid plot for activations for F1/siamese
%     \item Inertia plot
%     % \item (If space, 
% \end{itemize}

\begin{figure}[t]
  \centering
  \includegraphics[width=0.8\linewidth]{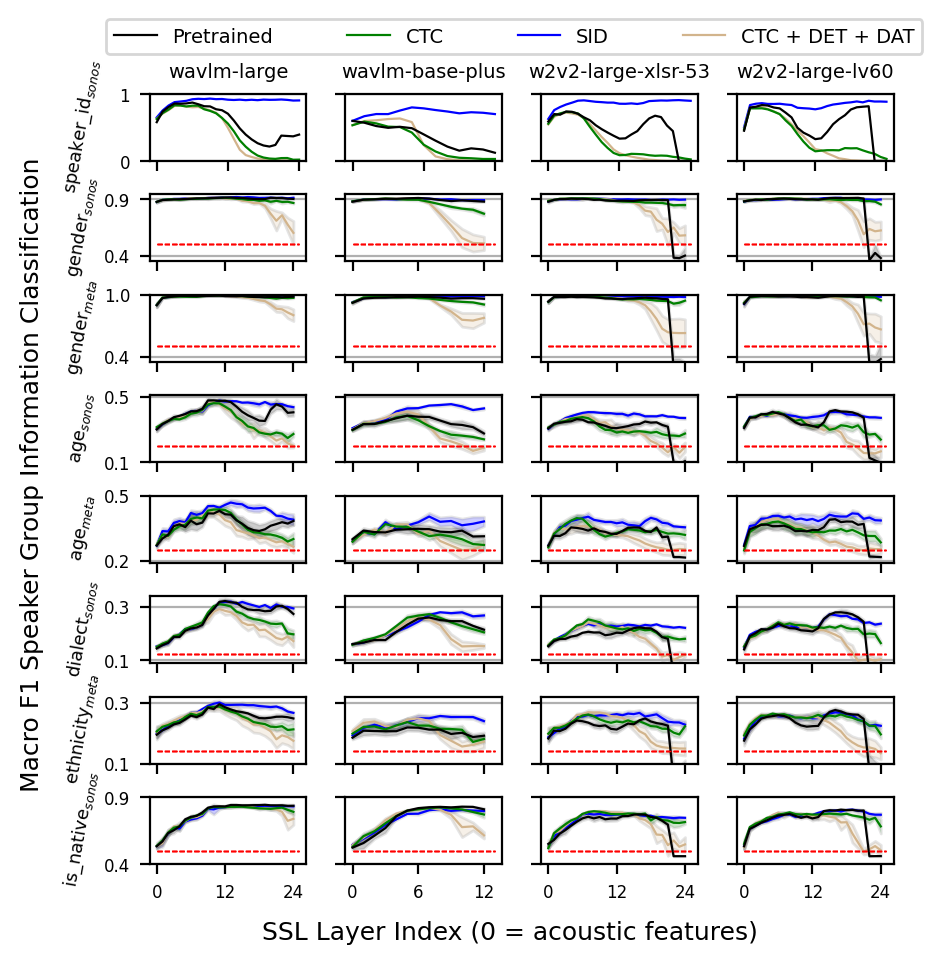}
  \caption{SGI captured by S3Ms, measured using \textit{linear probes}, as described in Equation \ref{eq:lin_reg}.}

  \label{fig:by_layer_line_graph_both}
\end{figure}

% and \ref{fig:layer_flow_base_models_ls_pretrain}
We illustrate the macro F1 probing accuracy for each SGC probe and model in Figure \ref{fig:by_layer_line_graph_both}. Every model we tested achieves greater than random performance for most SGCs on both datasets, though the SGs in \meta{} were less well detected. Ethnicity is barely detected by some models, and not detected by others. Socio-economic background was not consistently detectable by our probes for any models, and is not pictured in Figure \ref{fig:by_layer_line_graph_both}.We note the vastly superior performance of the SID task vs any individual SGI probing task. Despite having over 1000 classes (vs a handful for the SGI tasks), our probes achieved almost perfect performance for SID, while they were nowhere near perfect for any SGC besides gender.

% Because they tend to be less performant than \textbf{large} models (as has also been established for many other speech tasks \cite{superb}), the rest of study will focus on large models only. \footnote{The only instance when the \textbf{base} outperformed \textbf{large} was for \brq{}, wherein earlier layers contain significantly more information on dialect and ethnicity. It is unclear why a base model would ever be more performant at any task than a large model, provided proper pretraining.}

\subsection{Impact of model pretraining on SG encoding}

% \begin{figure}[t]
%   \centering
%   \includegraphics[width=\linewidth]{img/by_layer_line_graph/layer_flow_model_list_do_diff.png}
%   \caption{Difference in SG modeling capacity between large and base models}
%   \label{fig:layer_flow_model_list_do_diff}
% \end{figure}

The pretrained model architectures (black) which captures the most SGI are \wavlm{}-large. This aligns with our hypothesis that its reconstructive loss would lend itself more to SG encoding than a contrastive loss such as employed by \wtvt{} models. Unsurprisingly, \wavlm{}-large captures more SGI than \wavlm{}-base for most categories. Multilingual \wtvt{} tended to capture about the same amount of SGI as \wtvt{} pretrained on English only, which suggests that multilingual pretraining by itself does not force the model to learn such features.

% \begin{figure}[t]
%   \centering
%   \includegraphics[width=\linewidth]{img/by_layer_line_graph_both_acc-test_model_ft_type.png}
%   \caption{SGI captured by S3Ms, measured using \textit{Siamese speaker verification}.}
%   \label{fig:by_layer_line_graph_sonos_siamese}
% \end{figure}

% \begin{figure}[t]
%   \centering
%   \includegraphics[width=\linewidth]{img/by_layer_line_graph/layer_flow_base_models_ls_pretrain.png}
%   \caption{SGI captured by \textbf{base} pretrained S3M models. Title: $\text{category}_{dataset} < 1/|\text{category}|$}
%   \label{fig:layer_flow_base_models_ls_pretrain}
% \end{figure}

%  However, some SGs are better implicitly modeled than others. Most models can separate between men and women nearly perfectly, and fairly well between age and socioeconomic groups. However, they are less able to distinguish between native and non-native speakers, as well as between different dialects and ethnicities.

\subsection{Finetuning for ASR/SID reveals two classes of SGCs}

For all model variants depicted in Figure \ref{fig:by_layer_line_graph_both}, we observe that (apart from gender) comparatively little SGI is captured in the earliest layers of most models. This increases until just before halfway, i.e. layer 10 for \largem{} and 5 for \basem{} models. In middle and later layers, pretrained models (black) tend to retain relatively consistent levels of SGI for most classes, though the final layers of \wtvt{} models break this pattern (behavior which was also observed in \cite{toyota_layerwise_interpretability}). 

However, in the finetuned models, there are \textbf{two main behavior patterns} we observe in the \textit{middle and later layers}. Focusing on the models finetuned on vanilla ASR using CTC (green), we note some SGCs remain detectable at \textbf{relatively consistent levels} - these are dialect, ethnicity, and is\_native. The detection rate of these SGCs tends to continue increasing past the halfway layer, and subsequently either decreases only slightly or even increases (for is\_native), despite model finetuning. However, other SGCs become \textbf{considerably less detectable} in later layers - these are gender and age. In contrast, consider the models finetuned on the SID task (blue). SID finetuned models permit detection of gender and age (as well as speaker\_id) through later layers to a much greater extent than do the vanilla CTC models, and often greater than the pretrained models as well. However, SID finetuning does not improve detection of dialect, ethnicity, or is\_native (though it doesn't attenuate it either).

We hypothesize that this contrastive behavior is due to nature in which different SGs are manifested vocally. S3Ms are known to discard information during finetuning not relevant to the downstream task at hand \cite{forgetting_ssl}. Our observations thus support the notion that some SGCs are useful in ASR, while other SGCs are useful in SID. First, the SGCs that are useful for ASR are those which exhibit higher interclass semantic variance. The differences between different dialects', ethnicities', and native vs non-native speakers' speech tend to be prosodic/semantic - non-native speakers might insert pauses where native speakers would not pause; speakers of different dialects might use different phones to represent the same grapheme; speakers of different ethnicities might use different graphemes altogether \cite{sylvain_l2,aave_separate_models}. It is therefore intuitive that these SGCs remain relatively well detectable in later layers.

On the other hand, the SGCs which are associated with SID finetuning tend to be more phonetically variant. The primary distinguishing factors between the speech of different age groups and genders is more closely related to phonetic (women and children tend to have higher pitched voices, older people tend to have lower pitched voices, etc.). Previous work has shown that phonetic information encoded by S3Ms peaks towards the middle of models and then decreases, whereas semantic information peaks towards later layers \cite{toyota_layerwise_interpretability, toyota_layerwise_interpretability_2} - it is therefore not surprising that phonetically variant SGC detection peaks earlier for all model variants (though they retain their plateau with SID finetuning), while semantically variant SGC detection peaks slightly later.

\subsection{Ramifications of fairer ASR finetuning on SGI detection}

SGI detection on models finetuned with fairness enhancing ASR algorithms, i.e. CTC + DET (crimson) and CTC + DAT (orange) followed a similar pattern as we established in the previous section. Phonetic SGC was better detectable in CTC + DET than for vanilla CTC. On the other hand, models finetuned with CTC + DAT tend to slough off even more phonetically variant SGI in later layers than do vanilla CTC models - again this follows our intuition. However, CTC + DET models did not tend to encode semantically variant SGCs any better than vanilla ASR, nor did CTC + DAT models lose much by way of semantically variant SGI in later layers. This supports the notion that SID-based multi-task ASR finetuning should have the strongest effect on forcing performance invariance between phonetically variant SGCs, while it should be less effective on equalizing the performance between semantically variant SGCs. However, it is precisely the semantically variant classes that exhibit the greatest divergence in fairness for which DET and DAT finetuning appear to be least helpful \cite{sonos}. This motivates further study into alternative DET and DAT methods, such as were proposed in \cite{survey_on_accents_det_dat}, that use accent labels as targets, perhaps in addition to speaker IDs.

\begin{figure}[t]
  \centering
  \includegraphics[width=\linewidth]{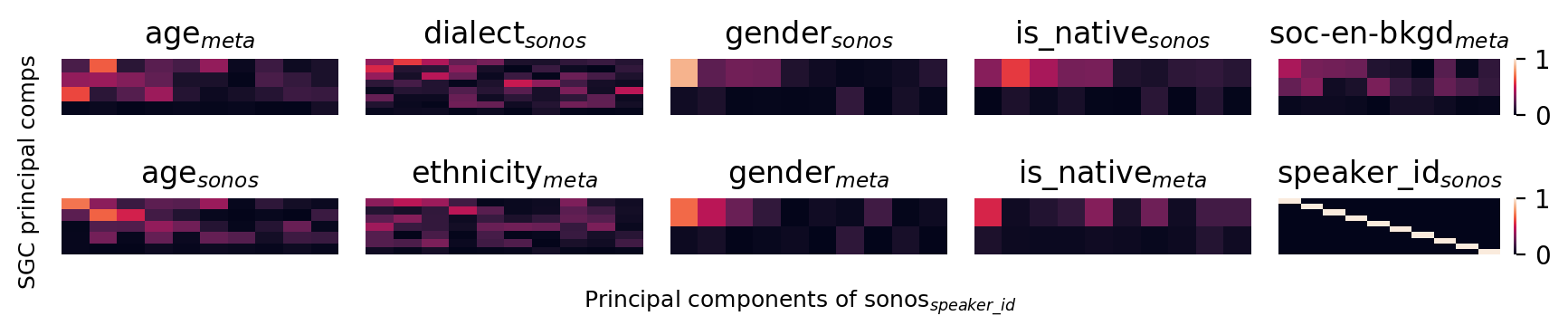}
  \caption{Cosine similarity between principal components of each SGC compared with principal components of the speaker centroid matrix, for layer 10 (i.e. layer with DET head) of \wavlm{}-large finetuned on CTC + DET + DAT.}
  \label{fig:cos_sim_centroids_layer_10}
\end{figure}

\begin{figure}[t]
  \centering
  \includegraphics[width=\linewidth]{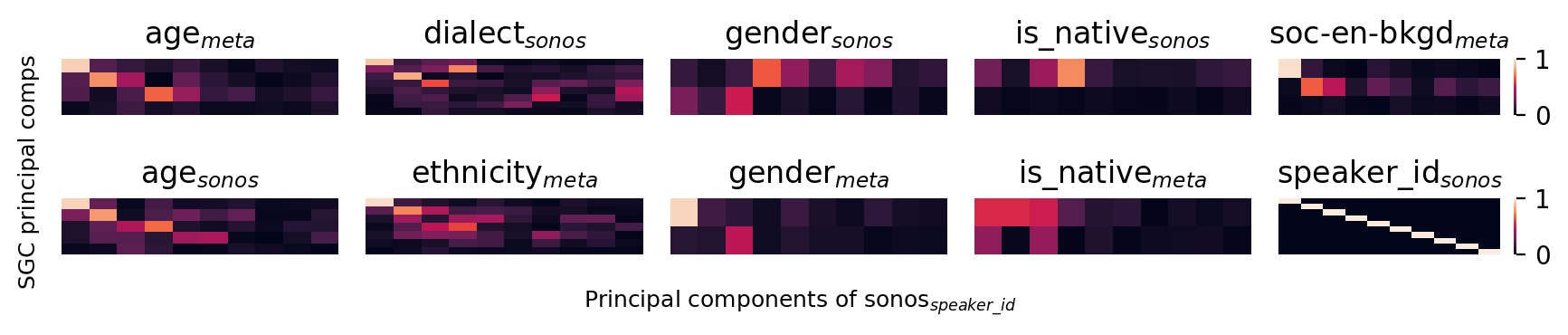}
  \caption{Cosine similarity between principal components of each SGC compared with principal components of the speaker centroid matrix, for layer 21 (i.e. layer with DAT head) of \wavlm{}-large finetuned on CTC + DET + DAT.}
  \label{fig:cos_sim_centroids_layer_21}
\end{figure}

% When models are finetuned on \textbf{ASR}, they tend lose some - in some cases up to 40\% - of their ability to recognize many SGs, as depicted in Figure \ref{fig:layer_flow_large_models_ft_comp_diff}\footnote{Strangely, the multilingual model improves in almost all categories. This could possibly explained by the model's focusing completely on English and thus allowing itself to discard information pertinent to multilingual modeling.}. The difference is particularly noticeable in the final few layers, likely seeing as how these are closest to the source of the gradient. This appears to be an example of the models forgetting what they had learned during their pretraining objectives in favor of focusing on ASR, a phenomenon that has been studied previously 

%Both datasets' gender recognition is affected the least, whereas dialect, ethnicity, and age categories are at least ten percent worse for most models. 

% \newpage

\newpage
\section{Ablation study on specific embedding dimensions}

\vspace{-0.4cm}

\begin{multicols}{2}

\subsection{SGI is distributed across frames}

\begin{figure}[H]
  \centering
  \includegraphics[width=0.95\linewidth]{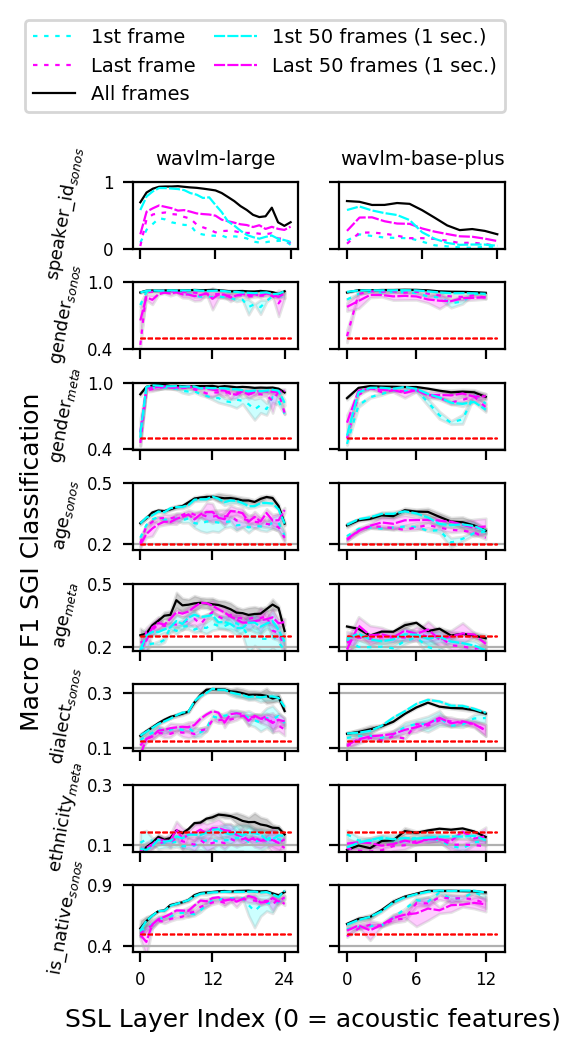}
  \caption{Earlier frames of a pretrained S3M capture more SGI than later frames, though the entire frame captures most of all.}
  \label{fig:Varying_pooling_window_impacts_speaker_group_recognition_ability_three_cat}
\end{figure}

% \vspace*{-1cm}
In Figure \ref{fig:Varying_pooling_window_impacts_speaker_group_recognition_ability_three_cat}, we observe the effect of pooling over different subsequences of the latent embeddings. We first note, not surprisingly, that pooling over 50 frames (long dashes) allows for far better SGI detection than single frames (short dashes). Previous work in \cite{silence_ssl_hubert} showed that the final frames of an encoded utterance contain most speaker ID information than earlier frames; however, we observe the opposite of this. For some SGCs (like gender), the difference is mostly negligible in comparing the pooled first second of audio and the final second of audio. However, for all other SGCs (which are more difficult to detect), pooling over the first 50 frames led to much greater SGI detection than the final 50 frames, a result more consistent with \cite{pasad_word_id_s3m} (though their analysis focused on word embeddings). The significantly superior performance of earlier layers could in part be due to the smart-speaker command format of our data - the regularity of the wake phrase at the start of each audio might provide a stable basis for SGI encoding.
\vspace*{\fill}
\columnbreak

\end{multicols}

%The self-attention mechanism of transformers supports the notion that global information might be present in very single frame, while some studies have shown some frame sequences to contain more SID-related information than others \cite{silence_S3M_hubert}.

\subsection{Different SGCs are encoded across different latent dimensions}

% \begin{figure}[t]
%   \centering
%   \includegraphics[width=\linewidth]{img/crv_step_graph_by_layer.png}
%   \caption{Cumulative residual variance (CRV) of the x-axis SGC with respect to the y-axis SGC. Lower values imply that more of the variance of the x-axis SGC is explained by the y-axis SGC.}
%   \label{fig:crv_step_graph}
% \end{figure}

\cite{orthogonality_first} show that SID information and phonetic information are stored along orthogonal dimensions within latent embeddings of S3Ms. They show this by comparing the principal components of the centroid matrix for each class, in their case phonemes vs speaker IDs. The centroid matrix is composed of the centroid (i.e. the mean over all samples) for each class (i.e. for gender that would mean two classes). The principal components of this matrix (measured by PCA) are the directions along which there is maximum variance between centroids of that SGC. If principal components of centroid matrices $M_{SGC_1}, M_{SGC_2}$ are orthogonal (i.e. cosine similarity close to 0), that implies that the information encapsulated by $SGC_1$ and $SGC_2$ are stored along orthogonal dimensions. 

%A subsequent work suggested the cumulative residual variance (CRV) metric, which measures the extent to which the principal components of one SGC are orthogonal to those of another SGC. The higher the CRV, the more orthogonal the centroid matrices. (For a more detailed explanation, see \cite{crv_pca} Section 3.2). 

%Our analysis is slightly different - we investigate whether certain (groups of) neurons specialize in encoding certain SGCs. To achieve this, we consider the weights learned by the two-tiered linear Siamese SGI discrimination models (see Equation \ref{eq:siamese} in Section \ref{sec:methods}). We first project our input features into a low-dimensional subspace, then measure similarity in that subspace (see Equation \ref{eq:cossim}). We can therefore think of these low-dimensional subspaces as representing each SGC in general - the similarity of those embeddings is what determines our prediction for SGC equality. The linear discriminator learns which input dimensions are useful to define this projected subspace; we can therefore look at which weights are highest, and compare these between different SGCs (see Equation \ref{eq:maxdim}. If two SGCs are modeled by the same neurons in our embedding, they will have high absolute valued weights in the linear projection from the latent transformer embedding to the low-dimensional SGC embedding. We can therefore compare these weight magnitudes between SGC classifiers to determine whether the same weights are used in modelling different SGCs.

In Figures \ref{fig:cos_sim_centroids_layer_10} and \ref{fig:cos_sim_centroids_layer_21}, we depict the cosine similarity ($S_C$) between pairs of principal components of SGCs. Higher values indicate correlated principal components. Particularly noteworthy are the values in the top left corner of each plot - these correspond to the covariance of the greatest principal components of SID with each SGC. Note the high degree of correlation between principal components for all SGCs at layer 10 (the DET layer); however, in layer 21 (the DAT layer), is\_native and and gender (for \sonos{}) correlate less with the first principal components of speaker\_id. This implies that DAT forces gender and is\_native \sonos{} to be modeled in an orthogonal dimension to speaker\_id, rather than render the layer 21 embeddings invariant to those SGCs\footnote{In Figure \ref{fig:by_layer_line_graph_both} we saw that CTC + DAT finetuning hampered the model's ability to detect speakers' gender. It is therefore puzzling that \sonos{} gender and speaker\_id are so uncorrelated.
}

\section{Outlook}

In this paper we provide a framework through which to evaluate S3Ms' ability to detect SGs. We show that a model's pretraining has a strong effect on how much SGI it captures. We also show that forcing an ASR-finetuned S3M to be SID invariant in its final layers renders the model blind towards more phonetically variant SGCs like gender and age; this forms a theoretical foundation on why DAT works for phonetic SGCs. However, semantically variant SGI (such as dialect or is\_native) tends to be retained in later layers in pretrained and finetuned models; a DAT head forcing speaker invariance has negligible effect on semantic SGI encoding. In our ablation study we showed that forcing speaker invariance causes certain SGCs to be encoded on orthogonal axes to speaker\_id. This further supports the hypothesis that DAT is less than optimally effective in suppressing those SGCs in later layers. That said, the extent to which SGI is necessary for high downstream task performance demands further study. Based on our results, we recommend experiment with variants of DET/DAT combining multiple SGC classifiers to amplify specific types of SGI in specific model layers.

% \section{Introduction}
% Your content.

% \subsection{Subsection}
% One example of the reference~\cite{paper}.

% \section{Detailed Information}
% General information about the Springer's LNCS series and links to useful materials for the authors 
% are available from the following URL:~
% \url{https://www.springer.com/us/computer-science/lncs/conference-proceedings-guidelines}

% Details about how to typeset specific paper parts (like e.g. formulae, code excerpts, etc.) can be found in the
% following document:~
% \url{https://resource-cms.springernature.com/springer-cms/rest/v1/content/19242230/data/v10}

% Bibliography
\bibliographystyle{plainnat}
\bibliography{these}

\end{document}